\documentclass[11pt,a4paper]{article}
\usepackage[hyperref]{naaclhlt2019}
\usepackage{times}
\usepackage{latexsym}
\usepackage{graphicx}
\usepackage{amsmath}
\usepackage{amssymb}
\usepackage{algorithm}
\usepackage{algorithmic}
\usepackage{linguex}
\usepackage{tabu}
\usepackage{subcaption} 
\usepackage{makecell}
\usepackage{pbox}
\usepackage{adjustbox}
\usepackage[T1]{fontenc}
\usepackage{xcolor}
\usepackage{url}

\definecolor{mypink}{RGB}{219, 48, 122}

\aclfinalcopy 

\title{Beyond The Wall Street Journal: \\ Anchoring and Comparing Discourse Signals across Genres}

\author{Yang Liu \\
  Department of Linguistics \\
  Georgetown University \\
  {\tt yl879@georgetown.edu}}

\date{}

\begin{document}
\maketitle
\begin{abstract}
  Recent research on discourse relations has found that they are cued not only by discourse markers (DMs) but also by other textual signals and that signaling information is indicative of genres. While several corpora exist with discourse relation signaling information such as the Penn Discourse Treebank (PDTB, \citealt{PrasadEtAl2008}) and the Rhetorical Structure Theory Signalling Corpus (RST-SC, \citealt{das2018rst}), they both annotate the Wall Street Journal (WSJ) section of the Penn Treebank (PTB, \citealt{MarcusSantoriniMarcinkiewicz1993}), which is limited to the news domain. Thus, this paper adapts the signal identification and anchoring scheme \cite{liu2019discourse} to three more genres, examines the distribution of signaling devices across relations and genres, and provides a taxonomy of indicative signals found in this dataset. 
\end{abstract}

\section{Introduction} \label{intro}

Sentences do not exist in isolation, and the meaning of a text or a conversation is not merely the sum of all the sentences involved: an informative text contains sentences whose meanings are relevant to each other rather than a random sequence of utterances. Moreover, some of the information in texts is not included in any one sentence but in their arrangement. Therefore, a high-level analysis of discourse and document structures is required in order to facilitate effective communication, which could benefit both linguistic research and NLP applications. For instance, an automatic discourse parser that successfully captures how sentences are connected in texts could serve tasks such as information extraction and text summarization.

A discourse is delineated in terms of relevance between textual elements. One of the ways to categorize such relevance is through \textit{coherence}, which refers to semantic or pragmatic linkages that hold between larger textual units such as \textsc{Cause, Contrast}, and \textsc{Elaboration} etc.  Moreover, there are certain linguistic devices that systematically signal certain discourse relations: some are generic signals across the board while others are indicative of particular relations in certain contexts. Consider the following example from the Georgetown University Multilayer (GUM) corpus \cite{zeldes2017gum},\footnote{The square brackets at the end of each example contain the document ID from which this example is extracted. Each ID consists of its genre type and one keyword assigned by the annotator at the beginning of the annotation task.} in which the two textual units connected by the DM \textit{but} form a \textsc{Contrast} relation, meaning that the contents of the two textual units are comparable yet not identical.

\ex. Related cross-cultural studies have resulted in insufficient statistical power, \textit{\textbf{but}} interesting trends (e.g., Nedwick, 2014 ). [academic\_implicature] \label{implicature}


However, the coordinating conjunction \textit{but} is also a frequent signal of another two relations that can express adversativity: \textsc{Concession} and \textsc{Antithesis}. \textsc{Concession} means that the writer acknowledges the claim presented in one textual unit but still claims the proposition presented in the other discourse unit while \textsc{Antithesis} dismisses the former claim in order to establish or reinforce the latter. In spite of the differences in their pragmatic functions, these three relations can all be frequently signaled by the coordinating conjunction \textit{but}: symmetrical \textsc{Contrast} as in \ref{implicature}, \textsc{Concession} as in \ref{implicature_concession}, and \textsc{Antithesis} as in \ref{chomsky_antithesis}. It is clear that \textit{but} is a generic signal here as it does not indicate strong associations with the relations it signals. 

\ex. This was a very difficult decision, \textit{\textbf{but}} one that was made with the American public in mind. [news\_nasa] \label{implicature_concession}

\ex. NATO had never rescinded it, \textit{\textbf{but}} they had and started some remilitarization. [interview\_chomsky] \label{chomsky_antithesis}

As suggested by \citet{taboada2003rhetorical}, some discourse signals are indicative of certain genres: they presented how to characterize appointment-scheduling dialogues using their rhetorical and thematic patterns as linguistic evidence and suggested that the rhetorical and the thematic analysis of their data can be interpreted functionally as indicative of this type of task-oriented conversation. Furthermore, the study of the classification of discourse signals can serve as valuable evidence to investigate their role in discourse as well as the relations they signal. 

One limitation of the RST Signalling Corpus is that no information about the location of signaling devices was provided. As a result, \citet{liu2019discourse} presented an annotation effort to anchor discourse signals for both elementary and  complex units on a small set of documents in RST-SC (see Section \ref{anchoring} for details). The present study addresses methodological limitations in the annotation process as well as annotating more data in more genres in order to investigate the distribution of signals across relations and genres and to provide both quantitative and qualitative analyses on signal tokens.

\section{Background}

\textbf{Rhetorical Structure Theory} (RST, \citealt{mann1988rhetorical}) is a well-known theoretical framework that extensively investigates discourse relations and is adopted by \citet{das2017signalling} and the present study. RST is a functional theory of text organization that identifies hierarchical structure in text. The original goals of RST were discourse analysis and proposing a model for text generation; however, due to its popularity, it has been applied to several other areas such as theoretical linguistics, psycholinguistics, and computational linguistics \citep{taboada2006applications}.  

RST identifies hierarchical structure and nuclearity in text, which categorizes relations into two structural types: \textsc{Nucleus-Satellite} and \textsc{Multinuclear}. The \textsc{Nucleus-Satellite} structure reflects a hypotactic relation whereas the \textsc{Multinuclear} structure is a paratactic relation \cite{taboada2013annotation}. The inventory of relations used in the RST framework varies widely, and therefore the number of relations in an RST taxonomy is not fixed. The original set of relations defined by \citet{mann1988rhetorical} included 23 relations. Moreover, RST identifies textual units as Elementary Discourse Units (EDUs), which are non-overlapping, contiguous spans of text that relate to other EDUs \cite{zeldes2017gum}. EDUs can also form hierarchical groups known as complex discourse units.

\subsection{Relation Signaling} \label{rs}

\begin{figure}
\centering
\includegraphics[width=75mm]{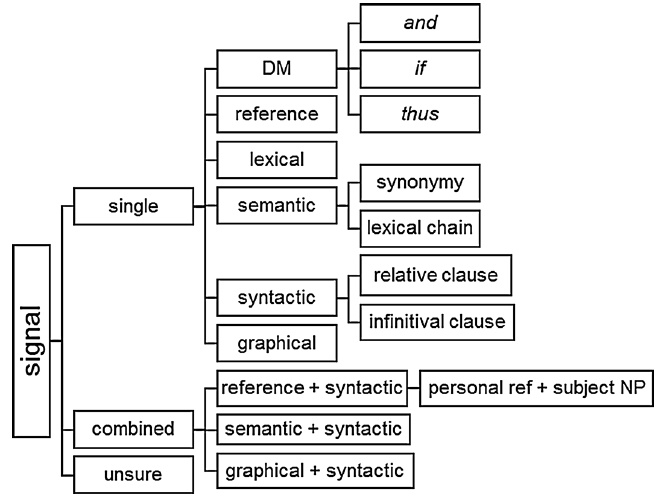}
\caption{Taxonomy of Signals in RST-SC (Fragment).}
\label{taxonomy}
\end{figure}

When it comes to relation signaling, the first question to ask is what a signal is. In general, signals are the means by which humans identify the realization of discourse relations. The most typical signal type is DMs (e.g.~`although') as they provide explicit and direct linking information between clauses and sentences. As mentioned in Section \ref{intro}, the lexicalized discourse relation annotations in PDTB have led to the discovery of a wide range of expressions called \textsc{Alternative Lexicalizations (\textit{AltLex})} \cite{prasad2010realization}. RST-SC provides a hierarchical taxonomy of discourse signals beyond DMs (see Figure \ref{taxonomy} for an illustration, reproduced from \citet[p.752]{das2017signalling}. 

Intuitively, DMs are the most obvious linguistic means of signaling discourse relations, and therefore extensive research has been done on DMs. Nevertheless, focusing merely on DMs is inadequate as they can only account for a small number of relations in discourse. To be specific, \citet{das2017signalling} reported that among all the 19,847 signaled relations (92.74\%) in RST-SC (i.e.~385 documents and all 21,400 annotated relations), relations exclusively signaled by DMs only account for 10.65\% whereas 74.54\% of the relations are exclusively signaled by other signals, corresponding to the types they proposed.

\subsection{The Signal Anchoring Mechanism}
\label{anchoring}

As mentioned in Section \ref{intro}, RST-SC does not provide information about the location of discourse signals. Thus, \citet{liu2019discourse} presented an annotation effort to anchor signal tokens in the text, with six categories being annotated. Their results showed that with 11 documents and 4,732 tokens, 923 instances of signal types/subtypes were anchored in the text, which accounted for over 92\% of discourse signals, with the signal type \textit{semantic} representing the most cases (41.7\% of signaling anchors) whereas discourse relations anchored by DMs were only about 8.5\% of anchor tokens in this study, unveiling the value of signal identification and anchoring. 

\subsection{Neural Modeling for Signal Detection}

\citet{Zeldes2018} trained a Recurrent Neural Network (RNN) model for the task of relation classification, and then latent associations in the network were inspected to detect signals. 
It is relatively easy to capture DMs such as `then' or a relative pronoun `which' signaling an \textsc{Elaboration}. The challenge is to figure out what features the network needs to know about beyond just word forms such as meaningful repetitions and variable syntactic constructions. With the human annotated data from the current project, it is hoped that more insights into these aspects can help us engineer meaningful features in order to build a more informative computational model.




\section{Methodology}


\textbf{Corpus}. The main goal of this project is to anchor and compare discourse signals across genres, which makes the Georgetown University Multilayer (GUM) corpus the optimal candidate, in that it consists of eight genres including interviews, news stories, travel guides, how-to guides, academic papers, biographies, fiction, and forum discussions. Each document is annotated with different annotation layers including but not limited to dependency (\texttt{dep}), coreference (\texttt{ref}), and rhetorical structures (\texttt{rst}). For the purpose of this study, the \texttt{rst} layer is used as it includes annotation on discourse relations, and signaling information will be anchored to it in order to produce a new layer of annotation. However, it is worth noting that other annotation layers are great resources to delve into discourse signals on other levels. 

\begin{figure}
\centering
\includegraphics[width=80mm]{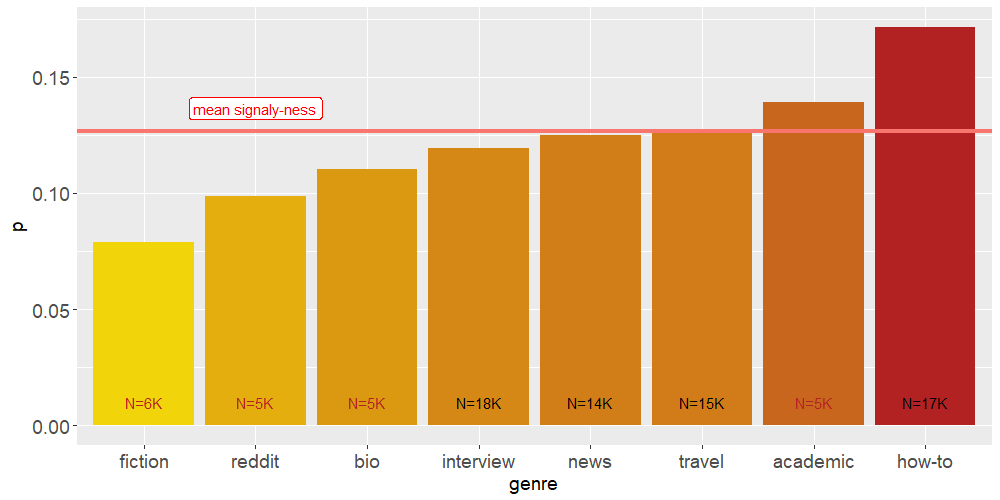}
\caption{A Visualization of How Strongly Each Genre Signals in the GUM Corpus.}
\label{gurt}
\end{figure}

Moreover, due to time limitations and the fact that this is the first attempt to apply the taxonomy of signals and the annotation scheme to other genres outside RST-DT's newswire texts, four out of eight genres in the GUM corpus were selected: \textit{academic}, \textit{how-to guides}, \textit{interviews}, and \textit{news}, which include a collection of 12 documents annotated for discourse relations. The rationale for choosing these genres is that according to \citet{Zeldes2018}'s neural approach to discourse signal prediction on the GUM corpus, how-to guides and academic articles in the GUM corpus signal most strongly, with interviews and news articles slightly below the average and fiction and reddit texts the least signaled, as shown in Figure \ref{gurt} (reproduced from \citet[p.19]{zeldes2018neural}). It is believed that the selection of these four genres is a good starting point of the topic under discussion.

\textbf{Annotation Tool}. One of the reasons that caused low inter-annotator agreement (IAA) in \citet{liu2019discourse} is the inefficient and error prone annotation tools they used: no designated tools were available for the signal anchoring task at the time. We therefore developed a better tool tailored to the purpose of the annotation task. It is built over an interface offering full RST editing capabilities called rstWeb \cite{Zeldes2016} and provides mechanisms for viewing and editing signals \cite{Gessler2019}. 

\textbf{Annotation Reliability}. In order to evaluate the reliability of the scheme, a revised inter-annotator agreement study was conducted using the same metric and with the new interface on three documents from RST-SC, containing 506 tokens with just over 90 signals. Specifically, agreement is measured based on token spans. That is, for each token, whether the two annotators agree it is signaled or not. The results demonstrate an improvement in Kappa, 0.77 as opposed to the previous Kappa 0.52 in \citet{liu2019discourse}. 

\textbf{Taxonomy of Discourse Signals}. The most crucial task in signaling annotation is the selection of signal types. The taxonomy of discourse signals used in this project is adapted from that of \citet{das2017signalling}, with additional types and subtypes to better suit other genres. Two new types and four new subtypes of the existing types are proposed: the two new types are \textit{Visual} and \textit{Textual} in which the subtype of the former is \textit{Image} and the subtypes of the latter are \textit{Title, Date,} and \textit{Attribution}. The three new subtypes are \textit{Modality} under the type \textit{Morphological} and \textit{Academic article layout, Interview layout} and \textit{Instructional text layout} under the type \textit{Genre}.

\textbf{Signal Anchoring Example}. Semantic features have several subtypes, with \textit{lexical chain} being the most common one. Lexical chains are annotated for words with the same lemma or words or phrases that are semantically related. Another characteristic of lexical chains is that words or phrases annotated as lexical chains are open to different syntactic categories. For instance, the following example shows that the relation \textsc{Restatement} is signaled by a \textit{lexical chain} item corresponding to the phrase \textit{a lot of} in the nucleus span and \textit{quantity} in the satellite span respectively. 

\ex. [They compensate for this by creating the impression that they have \textit{\textbf{a lot of}} friends --]\textsubscript{N} [they have a `\textit{\textbf{quantity}}, not quality' mentality.]\textsubscript{S} [whow\_arrogant] \label{lc}

\section{Results \& Analysis}

This pilot study annotated 12 documents with 11,145 tokens across four different genres selected from the GUM corpus. \textit{Academic articles, how-to guides}, and \textit{news} are written texts while \textit{interview} is spoken language. Generally speaking, all 20 relations used in the GUM corpus are signaled and anchored. However, this does not mean that all occurrences of these relations are signaled and anchored. There are several signaled but unanchored relations, as shown in Table \ref{unanchored}. In particular, the 5 unsignaled instances of the relation \textsc{Joint} result from the design of the annotation scheme (see Section \ref{scheme} for details). Additionally, the unanchored signal types and subtypes are usually associated with high-level discourse relations and usually correspond to genre features such as \textit{interview layout} in interviews where the conversation is constructed as a question-answer scheme and thus rarely anchored to tokens. 

With regard to the distribution of the signal types found in these 12 documents, the 16 distinct signal types amounted to 1263 signal instances, as shown in Table \ref{overviewresults}. There are only 204 instances of DMs out of all 1263 annotated signal instances (16.15\%) as opposed to 1059 instances (83.85\%) of other signal types. In RST-SC, DM accounts for 13.34\% of the annotated signal instances as opposed to 81.36\%\footnote{This result excludes the class \textit{Unsure} used in RST-SC.} of other signal types \cite{das2017signalling}. The last column in Table \ref{overviewresults} shows how the distribution of each signal type found in this dataset compares to RST-SC. The reason why the last column does not sum to 100\% is that not all the signal types found in RST-SC are present in this study such as the combined signal type \textit{Graphical + syntactic}. And since \textit{Textual} and \textit{Visual} are first proposed in this study, no results can be found in RST-SC, and the category \textit{Unsure} used in RST-SC is excluded from this project.

\subsection{Distribution of Signals across Relations}
\label{distribution-rels}

Table \ref{relation_based} provides the distribution of discourse signals regarding the relations they signal. The first column lists all the relations used in the GUM corpus. The second column shows the number of signal instances associated with each relation. The third and fourth columns list the most signaled and anchored type and subtype respectively.

The results show a very strong dichotomy of relations signaled by DMs and semantic-related signals: while DMs are the most frequent signals for five of the relations -- \textsc{Condition, Concession, Antithesis, Cause}, and \textsc{Circumstance}, the rest of the relations are all most frequently signaled by the type \textit{Semantic} or \textit{Lexical}, which, broadly speaking, are all associated with open-class words as opposed to functional words or phrases. Furthermore, the type \textit{Lexical} and its subtype \textit{indicative word} seem to be indicative of \textsc{Justify} and \textsc{Evaluation}. This makes sense due to the nature of the relations, which requires writers' or speakers' opinions or inclinations for the subject under discussion, which are usually expressed through positive or negative adjectives (e.g.~\textit{serious, outstanding, disappointed}) and other syntactic categories such as nouns/noun phrases (e.g.~\textit{legacy, excitement, an unending war}) and verb phrases (e.g.~\textit{make sure, stand for}). Likewise, words like \textit{Tips, Steps}, and \textit{Warnings} are indicative items to address communicative needs, which is specific to a genre, in this case, the how-to guides. It is also worth pointing out that \textsc{Evaluation} is the only discourse relation that is not signaled by any DMs in this dataset. 

\begin{table}[]
\small
\centering
\begin{tabular}{|c|c|c|}
\hline
\textit{\textbf{\begin{tabular}[c]{@{}c@{}}unanchored\\ relations\end{tabular}}} & \multicolumn{1}{l|}{\textit{\textbf{frequency}}} & \textit{\textbf{\begin{tabular}[c]{@{}c@{}}percentage (\%)\end{tabular}}} \\ \hline
\textsc{\textbf{Preparation}} & \color{red}28 & 22.2 \\ \hline
\textsc{\textbf{Solutionhood}} & 11 & \color{red}32.35 \\ \hline
\textsc{Joint} & 5 & 1.92 \\ \hline
\textsc{Background} & 3 & 2.68 \\ \hline
\textsc{Cause} & 1 & 4 \\ \hline
\textsc{Evidence} & 1 & 4.2 \\ \hline
\textsc{Motivation} & 1 & 4.76 \\ \hline
\end{tabular}
\caption{Distribution of Unanchored Relations.}
\label{unanchored}
\end{table}

\begin{table}[]
\begin{adjustbox}{width=\columnwidth,center}
\begin{tabular}{|c|c|c|c|}
\hline
\textit{\textbf{signal\_type}} & \textit{\textbf{frequency}} & \textit{\textbf{\makecell{percentage \\ (\%)}}} &
\textit{\textbf{\makecell{RST-SC \\ (\%)}}}\\ \hline
Semantic & 563 & \color{red}44.58 & \color{blue}24.8 \\ \hline
DM & 204 & 16.15 & 13.34 \\ \hline
Lexical & 156 & \color{red}12.35 & \color{blue}3.89 \\ \hline
Reference & 71 & 5.62 & 2.00 \\ \hline
Semantic + syntactic & 51 & 4.04 & 7.36 \\ \hline
Graphical & 46 & 3.64 & 3.46 \\ \hline
Syntactic & 44 & \color{blue}3.48 & \color{red}29.77 \\ \hline
Genre & 30 & 2.38 & 3.22 \\ \hline
Morphological & 26 & 2.06 & 1.07 \\ \hline
Syntactic + semantic & 25 & 1.98 & 1.40 \\ \hline
Textual & 24 & 1.90 & N/A \\ \hline
Numerical & 8 & 0.63 & 0.09 \\ \hline
Visual & 7 & 0.55 & N/A \\ \hline
Reference + syntactic & 3 & 0.24 & 1.86 \\ \hline
Lexical + syntactic & 3 & 0.24 & 0.41 \\ \hline
Syntactic + positional & 2 & 0.16 & 0.23 \\ \hline
Total & 1263 & 100.00 & 92.9 \\ \hline
\end{tabular}
\end{adjustbox}
\caption{Distribution of Signal Types and its Comparison to RST-SC.}
\label{overviewresults}
\end{table}

\begin{table}[]
\begin{adjustbox}{width=\columnwidth,center}
\begin{tabular}{|c|c|l|l|}
\hline
\textit{\textbf{\begin{tabular}[c]{@{}c@{}}signaled \\ relations\end{tabular}}} & \textit{\textbf{\begin{tabular}[c]{@{}c@{}}signal \\ instances\end{tabular}}} & \multicolumn{1}{c|}{\textit{\textbf{\begin{tabular}[c]{@{}c@{}}signal\\ type\end{tabular}}}} & \multicolumn{1}{c|}{\textit{\textbf{\begin{tabular}[c]{@{}c@{}}signal\\ subtype\end{tabular}}}} \\ \hline
\textsc{Joint} & 260 & Semantic (147) & lexical chain (96) \\ \hline
\textsc{Elaboration} & 243 & Semantic (140) &    lexical chain (96) \\ \hline
\textsc{Preparation} & 129 & Semantic (54) &    lexical chain (30) \\ \hline
\textsc{Background} & 112 & Semantic (62) &   lexical chain (42) \\ \hline
\textsc{Contrast} & 68 & Semantic (39) &    lexical chain (31) \\ \hline
\textsc{Restatement} & 60 & Semantic (34) &    lexical chain (28) \\ \hline
\textsc{\textbf{Concession}} & 49 & \color{red}DM (23) & \color{red}DM (23) \\ \hline
\textsc{\textbf{Justify}} & 49 & Lexical (25) & \color{red}indicative word (23) \\ \hline
\textsc{\textbf{Evaluation}} & 42 & Lexical (31) & \color{red}indicative word (31) \\ \hline
\textsc{Solutionhood} & 34 & Semantic (12) &    lexical chain (5) \\ \hline
\textsc{\textbf{Condition}} & 31 & \color{red}DM (25) & \color{red}DM (25) \\ \hline
\textsc{Antithesis} & 31 & DM (12) & DM (12) \\ \hline
\textsc{Sequence} & 26 & Semantic (7) & lexical chain (6) \\ \hline
\textsc{Cause} & 25 & DM (12) & DM (12) \\ \hline
\textsc{Evidence} & 24 & Semantic (8) & lexical chain (7) \\ \hline
\textsc{Result} & 21 & Semantic (8) & lexical chain (7) \\ \hline
\textsc{Motivation} & 21 & Semantic (8) & lexical chain (7) \\ \hline
\textsc{\textbf{Purpose}} & 21 & \color{red}Syntactic (9) & \color{red}infinitival clause (7) \\ \hline
\textsc{Circumstance} & 20 & DM (11) & DM (11) \\ \hline
\end{tabular}
\end{adjustbox}
\caption{Distribution of Most Common Signals across Relations. }
\label{relation_based}
\end{table}

Even though some relations are frequently signaled by DMs such as \textsc{Condition} and \textsc{Antithesis}, most of the signals are highly lexicalized and indicative of the relations they indicate. For instance, signal tokens associated with the relation \textsc{Restatement} tend to be the repetition or paraphrase of the token(s). Likewise, most of the tokens associated with \textsc{Evaluation} are strong positive or negative expressions. As for \textsc{Sequence}, in addition to the indicative tokens such as \textit{First \& Second} and temporal expressions such as \textit{later}, an indicative word pair such as \textit{stop \& update} can also suggest sequential relationship. More interestingly, world knowledge such as the order of the presidents of the United States (e.g.~that Bush served as the president of the United States before Obama) is also a indicative signal for \textsc{Sequence}.

\begin{table*}[]
\centering
\begin{tabular}{|c|l|}
\hline
\textit{\textbf{relations}} & \multicolumn{1}{c|}{\textit{\textbf{examples of anchored tokens}}} \\ \hline
\textsc{Joint} & ; (16), \textbf{and} (15), also (10), \textit{Professor Eastman} (3), he (3), they (2)\\ \hline
\textsc{Elaboration} & \makecell{Image (6), based on (3), -- (3), \textit{NASA} (3), \textit{IE6} (3), More specifically (2), \\ Additionally (2), also (2), they (2), it (2), \textit{Professor Chomsky} (2)} \\ \hline
\textsc{Preparation} & : (6), How to (2), Know (2), Steps (2), Getting (2) \\ \hline
\textsc{Background} & Therefore, Indeed, build on, previous, \textit{Bob McDonnell}, Looking back \\ \hline
\textsc{Contrast} & \makecell{\textbf{but} (9)/\textbf{But} (4), \textbf{or} (2), Plastic-type twist ties \& paper-type twist ties, \\ in 2009 \& today, deteriorate \& hold up, however, bad \& nice, yet} \\ \hline
\textsc{Restatement} & \makecell{They \& they (2), \textit{NATO} (2), In other words, realistic \& real, \textbf{and}, \\ rehashed \& retell, it means that, \textit{Microsoft \& Microsoft}} \\ \hline
\textsc{Concession} & \makecell{\textbf{but} (10), However (3), The problem is (2), though (2), at least, While, \\ It is (also) possible that, however, best \& okay, Albeit, despite, if, still} \\ \hline
\textsc{Justify} & \makecell{because (2), an affront \& disappointed deeply, excitement, share, \\ \textit{the straps}, The logic is that, any reason, \textbf{so}, \textbf{since}, confirm, inspire} \\ \hline
\textsc{Evaluation} & \makecell{very serious, nationally representative, a frightening idea, a true win, \\ an important addition, issue, This study \& It, misguided, pain} \\ \hline
\textsc{Solutionhood} & \makecell{Well (2), arrogant, :, \textbf{So}, why, \textbf{and}, \textit{Darfur}, How, I think, Determine} \\ \hline
\textsc{Condition} & \makecell{If (12)/if(10), even if, unless, depends on, --, once, when, until} \\ \hline
\textsc{Antithesis} & \makecell{\textbf{but} (5)/\textbf{But}, instead of (2), In fact, counteract, won't, rather than, \\ \textbf{Or}, not, \textit{the Arabs}, however, better \& worst} \\ \hline
\textsc{Sequence} & \makecell{\textbf{and} (3), First \& Second, examined \& assessed, later, Bush \& Obama, \\initial, \textit{digital humanities}, A year later, stop \& update} \\ \hline
\textsc{Cause} & \makecell{because (3), suggests, due to, compensate for, \textbf{as}, \textbf{since}/\textbf{Since}, \\ \textit{arrogant people}, in turn, given, brain damage, as such} \\ \hline
\textsc{Evidence} & \makecell{( ) (2), see (2), According to, because, \textbf{as}, --, \textbf{and}, \textit{Arabs \& Turkey}, \\ Because of, \textit{discrimination}, biases, The report states that, Thus} \\ \hline
\textsc{Result} & \makecell{\textbf{so} (3), \textbf{and} (2), meaning (2), so that, capturing, thus, putting, \\ \textit{the $\chi$2 statistic}, make} \\ \hline
\textsc{Motivation} & \makecell{will (2), easier, \textit{the pockets}, All it takes is, \textbf{so}, last longer} \\ \hline
\textsc{Purpose} & \makecell{to (6)/To, in order to (3)/In order to (2), \textbf{so} (2), enable, The aim} \\ \hline
\textsc{Circumstance} & \makecell{when (4)/When (2), On March 13, Whether, \textbf{As}/\textbf{as}, With, \\ in his MIT office, the bigger \& the harsher} \\ \hline
\end{tabular}
\caption{Examples of Anchored Tokens across Relations. }
\label{anchoredtokens}
\end{table*}

\begin{table*}[]
\centering
\begin{tabular}{|c|l|}
\hline
\multicolumn{1}{|l|}{} & \multicolumn{1}{c|}{\textit{\textbf{examples of anchored tokens in different genres}}} \\ \hline
\textit{\textbf{academic}} & \makecell{\textit{discrimination} (16), \textbf{;} (11), and (8), \textbf{:} (5), \textbf{to} (5), but (5), also (5), \\ though (3), \textbf{hypothesized} (3), \textbf{based on} (3), \textbf{First} \& \textbf{Second} (3), \\  however (3), because (2), More specifically (2), in/In order to (2), as (2), \\ \textbf{( )} (2), \textbf{see} (2), when (2), \textbf{posited}, \textbf{expected}, capturing, \textbf{Albeit}} \\ \hline
\textit{\textbf{\begin{tabular}[c]{@{}c@{}}how-to \\ guides\end{tabular}}} & \makecell{but (10), If (9)/if(7), \textbf{;} (5), and (4), also (4), \textit{arrogant people} (9), \\
\textbf{How} (7), \textbf{:} (3), so (3), \textbf{--} (3), But (3), \textbf{Know} (3), \textbf{Steps} (2), Move, \\ Challenge, \textbf{Warnings}, In other words, Empty, Fasten, \textbf{Tips}, Wash} \\ \hline
\textit{\textbf{news}} & \makecell{\textit{IE6} (9), \textit{NASA} (5), and (4), but (4)/But (2), \textbf{Image} (4), \textbf{market} (4) \\ However (2), \textbf{the major source}, the Udvar-Hazy Center, in 2009} \\ \hline
\textit{\textbf{interviews}} & \makecell{\textit{Sarvis} (14), \textbf{What} (12), \textbf{Why} (11), and (8), \textit{Noam Chomsky} (8), but (5), \\ \textbf{Wikinews} (4), because (3), \textbf{interview} (2), -- (2), \textbf{Well} (2), So (2), Which} \\ \hline
\end{tabular}
\caption{Examples of Anchored Tokens across Genres.}
\label{genre-tokens}
\end{table*}

Another way of seeing these signals is to examine their associated tokens in texts, regardless of the signal types and subtypes. Table \ref{anchoredtokens} lists some representative, generic/ambiguous (in boldface), and coincidental (in italics) tokens that correspond to the relations they signal. Each item is delimited by a comma; the \textit{\&} symbol between tokens in one item means that this signal consists of a word pair in respective spans. The number in the parentheses is the count of that item attested in this project; if no number is indicated, then that token span only occurs once. The selection of these single-occurrence items is random in order to better reflect the relevance in contexts. For instance, lexical items like \textit{Professor Eastman} in \textsc{Joint}, \textit{NASA} in \textsc{Elaboration}, \textit{Bob McDonnell} in \textsc{Background}, and \textit{NATO} in \textsc{Restatement} appear to be coincidental because they are the topics or subjects being discussed in the articles. These results are parallel to the findings in \citet[p.180]{Zeldes2018}, which employed a frequency-based approach to show the most distinctive lexemes for some relations in GUM.

\subsection{Distribution of Signals across Genres}
\label{distribution-genres}

Table \ref{genre_based} shows the distribution of the signaled relations in different genres. Specifically, the number preceding the vertical line is the number of signals indicating the relation and the percentage following the vertical line is the corresponding proportional frequency. The label \textsc{N/A} suggests that no such relation is present in the sample from that genre. 

\begin{table}[]
\begin{adjustbox}{width=\columnwidth,center}
\begin{tabular}{|c|c|c|c|c|}
\hline
\textit{\textbf{relations}} & \textit{\textbf{academic}} & \textit{\textbf{how-to guides}} & \textit{\textbf{news}} & \textit{\textbf{interview}} \\ \hline
{ \textsc{Joint}} & { \color{red}65 | 23.13\%} & { 76 | 18.67\%} & { \color{red}65 | 25.39\%} & { \color{red}54 | 16.77\%} \\ \hline
{ \textsc{Elaboration}} & { 61 | 21.71\%} & { \color{red}79 | 19.41\%} & { 53 | 20.70\%} & { 50 | 15.53\%} \\ \hline
{ \textsc{Preparation}} & { 25 | 8.9\%} & { \color{purple}55 | 13.51} & { 15 | 5.86\%} & { 34 | 10.56\%} \\ \hline
{ \textsc{Background}} & { \color{purple}33 | 11.74\%} & { 24 | 5.9\%} & { 28 | 10.94\%} & { 27 | 8.39\%} \\ \hline
{ \textsc{Contrast}} & { 17 | 6.05\%} & { \color{purple}21 | 5.16\%} & { 19 | 7.42\%} & { 11 | 3.42\%} \\ \hline
{ \textsc{Restatement}} & { N/A} & { 20 | 4.91\%} & { 11 | 4.3\%} & { \color{purple}29 | 9.01\%} \\ \hline
{ \textsc{Concession}} & { \color{purple}17 | 6.05\%} & { 13 | 3.19\%} & { 10 | 3.91\%} & { 9 | 2.8\%} \\ \hline
{ \textsc{Justify}} & { 1 | 0.36\%} & { 11 | 2.7\%} & { 15 | 5.86\%} & { \color{purple}22 | 6.83\%} \\ \hline
{ \textsc{Evaluation}} & { 10 | 3.56\%} & { 12 | 2.95\%} & { 7 | 2.73\%} & { \color{purple}13 | 4.04\%} \\ \hline
{ \textsc{Solutionhood}} & { 2 | 0.71\%} & { 8 | 1.97\%} & { N/A} & { \color{purple}24 | 7.45\%} \\ \hline
{ \textsc{Condition}} & { N/A} & { \color{purple}25 | 6.14\%} & { 3 | 1.17\%} & { 3 | 0.93\%} \\ \hline
{ \textsc{Antithesis}} & { 3 | 1.07\%} & { 10 | 2.46\%} & { 1 | 0.39\%} & { \color{purple}17 | 5.28\%} \\ \hline
{ \textsc{Sequence}} & { \color{purple}12 | 4.27\%} & { 4 | 0.98} & { 5 | 1.95\%} & { 5 | 1.55\%} \\ \hline
{ \textsc{Cause}} & { 6 | 2.14\%} & { \color{purple}12 | 2.95} & { 6 | 2.34\%} & { 1 | 0.31\%} \\ \hline
{ \textsc{Evidence}} & { \color{purple}10 | 3.56\%} & { N/A} & { 5 | 1.95\%} & { 9 | 2.8\%} \\ \hline
{ \textsc{Result}} & { 3 | 1.07\%} & { 6 | 1.47\%} & { \color{purple}6 | 2.34\%} & { 6 | 1.86\%} \\ \hline
{ \textsc{Motivation}} & { N/A} & { \color{purple}21 | 5.16\%} & { N/A} & { N/A} \\ \hline
{ \textsc{Purpose}} & { \color{purple}14 | 4.98\%} & { 5 | 1.23\%} & { N/A} & { 2 | 0.62\%} \\ \hline
{ \textsc{Circumstance}} & { 2 | 0.36\%} & { 5 | 1.23\%} & { \color{purple}7 | 2.73\%} & { 6 | 1.86\%} \\ \hline
{ \textit{Total}} & { 281 | 100\%} & { \textbf{\color{red}407} | 100\%} & { \textbf{\color{blue}256} | 100\%} & { 322 | 100\%} \\ \hline
\end{tabular}
\end{adjustbox}
\caption{Distribution of Signaled Relations across Genres.}
\label{genre_based}
\end{table}

As can be seen from Table \ref{genre_based}, how-to guides involve the most signals (i.e.~407 instances), followed by interviews, academic articles, and news. It is surprising to see that news articles selected from the GUM corpus are not as frequently signaled as they are in RST-SC, which could be  attributed to two reasons. Firstly, the source data is different. The news articles from GUM are from Wikinews while the documents from RST-SC are Wall Street Journal articles. Secondly, RST-DT has finer-grained relations (i.e.~78 relations as opposed to the 20 relations used in GUM) and segmentation guidelines, thereby having more chances for signaled relations. Moreover, it is clear that \textsc{Joint} and \textsc{Elaboration} are the most frequently signaled relations in all four genres across the board, followed by \textsc{Preparation} in how-to guides and interviews or \textsc{Background} in academic articles and news, which is expected as these four relations all show high-level representations of discourse that involve more texts with more potential signals. 

Table \ref{genre-tokens} lists some signal tokens that are indicative of genre (in boldface) as well as generic and coincidental ones (in italics). The selection of these items follows the same criteria used in Section \ref{distribution-rels}. Even though DMs \textit{and} and \textit{but} are present in all four genres, no associations can be established between these DMs and the genres they appear in. Moreover, as can be seen from Table \ref{genre-tokens}, graphical features such as semicolons, colons, dashes, and parentheses play an important role in relation signaling. Although these punctuation marks do not seem to be indicative of any genres, academic articles tend to use them more as opposed to other genres. Although some words or phrases are highly frequent, such as \textit{discrimination} in academic articles, \textit{arrogant people} in how-to guides, \textit{IE6} in news, and \textit{Sarvis} in interviews, they just seem to be coincidental as they happen to be the subjects or topics being discussed in the articles. 

\textbf{Academic writing} is typically formal, making the annotation more straightforward. The results from this dataset suggest that academic articles contain signals with diverse categories. As shown in Table \ref{genre-tokens}, in addition to the typical DMs and some graphical features mentioned above, there are several lexical items that are very strong signals indicating the genre. For instance, the verb \textit{hypothesized} and its synonym \textit{posited} are indicative in that researchers and scholars tend to use them in their research papers to present their hypotheses. The phrase \textit{based on} is frequently used to elaborate on the subject matter. Furthermore, Table \ref{genre-tokens} also demonstrates that academic articles tend to use ordinal numbers such as \textit{First} and \textit{Second} to structure the text. Last but not least, the word \textit{Albeit} indicating the relation \textsc{Concession} seems to be an indicative signal of academic writing due to the register it is associated with. 

\textbf{How-to Guides} are the most signaled genre in this dataset. This is due to the fact that instructional texts are highly organized, and the cue phrases are usually obvious to identify. As shown in Table \ref{genre-tokens}, there are several indicative signal tokens such as the \textit{wh}-word \textit{How}, an essential element in instructional texts. Words like \textit{Steps}, \textit{Tips}, and \textit{Warnings} are strongly associated with the genre due to its communicative needs. Another distinct feature of how-to guides is the use of imperative clauses, which correspond to verbs whose first letter is capitalized (e.g.~\textit{Know, Empty, Fasten, Wash}), as instructional texts are about giving instructions on accomplishing certain tasks and imperative clauses are good at conveying such information in a straightforward way. 

\textbf{News articles}, like academic writing, are typically organized and structured. As briefly mentioned at the beginning of this section, news articles selected in this project are not as highly signaled as the news articles in RST-SC. In addition to the use of different source data, another reason is that RST-DT employs a finer-grained relation inventory and segmentation guidelines; as a result, certain information is lost. For instance, the relation \textsc{Attribution} is signaled 3,061 times out of 3070 occurrences (99.71\%) in RST-SC, corresponding to the type \textit{syntactic} and its subtype \textit{reported speech}, which does not occur in this dataset. However, we do have some indicative signal tokens such as \textit{market} and \textit{the major source}. 

\textbf{Interviews} are the most difficult genre to annotate in this project for two main reasons. Firstly, it is (partly) spoken language; as a result, they are not as organized as news or academic articles and harder to follow. Secondly, the layout of an interview is fundamentally different from the previous three written genres. For instance, the relation \textsc{Solutionhood} seems specific to interviews, and most of the signal instances remain unanchored (i.e.~11 instances), which is likely due to the fact that the question mark is ignored in the current annotation scheme. As can be seen from Table \ref{genre-tokens}, there are many \textit{wh-}words such as \textit{What} and \textit{Why}. These can be used towards identifying interviews in that they formulate the question-answer scheme. Moreover, interviewers and interviewees are also important constituents of an interview, which explains the high frequencies of the two interviewees \textit{Sarvis} and \textit{Noam Chomsky} and the interviewer \textit{Wikinews}. Another unique feature shown by the signals in this dataset is the use of spoken expressions such as \textit{Well} and \textit{So} when talking, which rarely appear in written texts.

\section{Discussion} \label{discussion}

\subsection{Annotation Scheme} \label{scheme}

 For syntactic signals, one of the questions worth exploring is which of these are actually attributable to sequences of tokens, and which are not. For example, sequences of auxiliaries or constructions like imperative clauses might be identifiable, but more implicit and variable syntactic constructions are not such as ellipsis.
 
 In addition, one of the objectives of the current project is to provide human annotated data in order to see how the results produced by machine learning techniques compare to humans' judgments. In particular, we are interested in whether or not contemporary neural models have a chance to identify the constructions that humans use to recognize discourse relations in text based on individual sequences of word embeddings, a language modeling technique that converts words into vectors of real numbers that are used as the input representation to a neural network model based on the idea that words that appear in similar environments should be represented as close in vector space. 

\begin{figure}
\centering
\includegraphics[width=80mm]{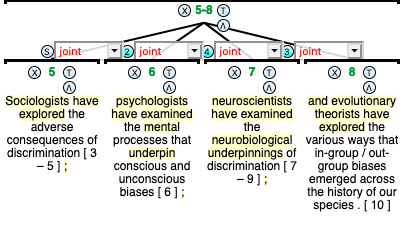}
\caption{A Visualization of a Multinuclear Relation.}
\label{multinuclear}
\end{figure}

 Another dilemma that generally came up during the discussion about signal anchoring was whether or not to mark the first constituent of a multinuclear relation. In Figure \ref{multinuclear}, four juxtaposed segments are linked together by the \textsc{Joint} relation, with associated signal tokens being highlighted. The first instance of \textsc{Joint} is left unsignaled/unmarked while the other instances of \textsc{Joint} are signaled. The rationale is that when presented with a parallelism, the reader only notices it from the second instance. As a result, signals are first looked for between the first two spans, and then between the second and the third. If there is no signal between the second and the third spans, then try to find signals in the first and the third spans. Because this is a multinuclear relation, transitivity does exist between spans. Moreover, the current approach is also supported by the fact that a multinuclear relation is often found in the structure like X, Y \textit{and} Z, in which the discourse marker \textit{and} is between the last two spans, and thus this \textit{and} is only annotated for the relation between the last two spans but not between the first two spans. However, the problem with this approach is that the original source for the parallelism cannot be located. 

\subsection{Distribution of Discourse Signals}

So far we have examined the distributions of signals across relations (Section \ref{distribution-rels}) and genres (Section \ref{distribution-genres}) respectively. Generally speaking, DMs are not only ambiguous but also inadequate as discourse signals; most signal tokens are open-class lexical items. More specifically, both perspectives have revealed the fact that some signals are highly indicative while others are generic or ambiguous. Thus, in order to obtain more valid discourse signals and parse discourse relations effectively, we need to develop models that take signals' surrounding contexts into account to disambiguate these signals. 

Based on the results found in this dataset regarding the indicative signals, they can be broadly categorized into three groups: \textit{\textbf{register-related}}, \textit{\textbf{communicative-need related}}, and \textit{\textbf{semantics-related}}. The first two are used to address genre specifications whereas the last one is used to address relation classification. Words like \textit{Albeit} are more likely to appear in academic papers than other genres due to the register they are associated with; words like \textit{Steps, Tips,} and \textit{Warnings} are more likely to appear in instructional texts due to the communication effect they intend to achieve. Semantics-related signals play a crucial role in classifying relations as the semantic associations between tokens are less ambiguous cues, thereby supplementing the inadequacy of DMs.

\subsection{Validity of Discourse Signals}

It is also worth pointing out that some tokens are frequent signals in several relations, which makes their use very ambiguous. For instance, the coordinating conjunction \textit{and} appears in \textsc{Joint, Restatement, Sequence}, and \textsc{Result} in this dataset. Similarly, the subordinating conjunctions \textit{since} and \textit{because} serve as signals of \textsc{Justify}, \textsc{Cause}, and \textsc{Evidence} in these 12 documents. These ambiguities would pose difficulties to the validity of discourse signals. As pointed out by \citet{Zeldes2018}, a word like \textit{and} is extremely ambiguous overall, since it appears very frequently in general, and is attested in all discourse functions. However, it is noted that some `and's are more useful as signals than others: adnominal `and' (example \ref{adnominal}) is usually less interesting than intersentential `and' (example \ref{intersentential}) and sentence initial `and' (example \ref{initial}). 

\ex. The owners, [William \textit{\textbf{and}} Margie Hammack], are luckier than any others.\footnote{This example is chosen from the RST-DT corpus \cite{CarlsonEtAl2003} for illustration due to the apposition. Note that the relation inventory also differs. } -- \textsc{Elaboration-Additional} \label{adnominal}

\ex. [Germany alone had virtually destroyed Russia, twice,]\textsubscript{n1} [\textit{\textbf{and}} Germany backed by a hostile military alliance, centered in the most phenomenal military power in history, that's a real threat.]\textsubscript{n2} -- \textsc{Joint} [interview\_chomsky] \label{intersentential}

\ex. [It arrests us.]\textsubscript{N} [\textit{\textbf{And}} then you say you won't commit a mistake, so you'll commit new mistakes. It doesn't matter.]\textsubscript{S} --\textsc{Antithesis} [interview\_peres] \label{initial}

Hence, it would be beneficial to develop computational models that score and rank signal words not just based on how proportionally often they occur with a relation, but also on how (un)ambiguous they are in contexts. In other words, if there are clues in the environment that can tell us to safely exclude some occurrences of a word, then those instances shouldn't be taken into consideration in measuring its `signalyness'.

\section{Conclusion} \label{conclusion}

The current study anchors discourse signals across several genres by adapting the hierarchical taxonomy of signals used in RST-SC. In this study, 12 documents with 11,145 tokens across four different genres selected from the GUM corpus are annotated for discourse signals. The taxonomy of signals used in this project is based on the one in RST-SC with additional types and subtypes proposed to better represent different genres. The results have shown that different relations and genres have their indicative signals in addition to generic ones, and the indicative signals can be characterized into three categories: register-related, communicative-need related, and semantics-related. 

The current study is limited to the \texttt{rst} annotation layer in GUM; it is worth investigating the linguistic representation of these signals through other layers of annotation in GUM such as coreference and bridging, which could be very useful resources constructing theoretical models of discourse. In addition, the current project provides a qualitative analysis on the validity of discourse signals by looking at the annotated signal tokens across relations and genres respectively, which provides insights into the disambiguation of generic signals and paves the way for designing a more informative mechanism to quantitatively measure the validity of discourse signals.  





\bibliography{naaclhlt2019}

\begin{thebibliography}{16}
\expandafter\ifx\csname natexlab\endcsname\relax\def\natexlab#1{#1}\fi

\bibitem[{Carlson et~al.(2003)Carlson, Marcu, and Okurowski}]{CarlsonEtAl2003}
Lynn Carlson, Daniel Marcu, and Mary~Ellen Okurowski. 2003.
\newblock Building a {D}iscourse-{T}agged {C}orpus in the {F}ramework of
  {R}hetorical {S}tructure {T}heory.
\newblock In \emph{Current and New Directions in Discourse and Dialogue}, Text,
  Speech and Language Technology 22, pages 85--112. Kluwer, Dordrecht.

\bibitem[{Das and Taboada(2017)}]{das2017signalling}
Debopam Das and Maite Taboada. 2017.
\newblock Signalling of {C}oherence {R}elations in {D}iscourse, {B}eyond
  {D}iscourse {M}arkers.
\newblock \emph{Discourse Processes}, pages 1--29.

\bibitem[{Das and Taboada(2018)}]{das2018rst}
Debopam Das and Maite Taboada. 2018.
\newblock R{ST} {S}ignalling {C}orpus: A {C}orpus of {S}ignals of {C}oherence
  {R}elations.
\newblock \emph{Language Resources and Evaluation}, 52(1):149--184.

\bibitem[{Gessler et~al.(2019)Gessler, Liu, and Zeldes}]{Gessler2019}
Luke Gessler, Yang Liu, and Amir Zeldes. 2019.
\newblock A {D}iscourse {S}ignal {A}nnotation {S}ystem for {RST} {T}rees.
\newblock In \emph{Proceedings of 7th Workshop on Discourse Relation Parsing
  and Treebanking (DISRPT) at NAACL-HLT}, Minneapolis, MN.
\newblock (To Appear).

\bibitem[{Liu and Zeldes(2019)}]{liu2019discourse}
Yang Liu and Amir Zeldes. 2019.
\newblock Discourse relations and signaling information: Anchoring discourse
  signals in {RST-DT}.
\newblock \emph{Proceedings of the Society for Computation in Linguistics},
  2(1):314--317.

\bibitem[{Mann and Thompson(1988)}]{mann1988rhetorical}
William~C Mann and Sandra~A Thompson. 1988.
\newblock Rhetorical {S}tructure {T}heory: Toward a {F}unctional {T}heory of
  {T}ext {O}rganization.
\newblock \emph{Text-Interdisciplinary Journal for the Study of Discourse},
  8(3):243--281.

\bibitem[{Marcus et~al.(1993)Marcus, Santorini, and
  Marcinkiewicz}]{MarcusSantoriniMarcinkiewicz1993}
Mitchell~P. Marcus, Beatrice Santorini, and Mary~Ann Marcinkiewicz. 1993.
\newblock Building a {L}arge {A}nnotated {C}orpus of {E}nglish: The {P}enn
  {T}reebank.
\newblock \emph{Special Issue on Using Large Corpora, Computational
  Linguistics}, 19(2):313--330.

\bibitem[{Prasad et~al.(2008)Prasad, Dinesh, Lee, Miltsakaki, Robaldo, Joshi,
  and Webber}]{PrasadEtAl2008}
Rashmi Prasad, Nikhil Dinesh, Alan Lee, Eleni Miltsakaki, Livio Robaldo,
  Aravind Joshi, and Bonnie Webber. 2008.
\newblock The {P}enn {D}iscourse {T}reebank 2.0.
\newblock In \emph{Proceedings of the 6th International Conference on Language
  Resources and Evaluation (LREC 2008)}, pages 2961--2968, Marrakesh, Morocco.

\bibitem[{Prasad et~al.(2010)Prasad, Joshi, and Webber}]{prasad2010realization}
Rashmi Prasad, Aravind Joshi, and Bonnie Webber. 2010.
\newblock Realization of {D}iscourse {R}elations by {O}ther {M}eans:
  {A}lternative {L}exicalizations.
\newblock In \emph{Proceedings of the 23rd International Conference on
  Computational Linguistics: Posters}, pages 1023--1031. Association for
  Computational Linguistics.

\bibitem[{Taboada and Das(2013)}]{taboada2013annotation}
Maite Taboada and Debopam Das. 2013.
\newblock Annotation upon {A}nnotation: Adding {S}ignalling {I}nformation to a
  {C}orpus of {D}iscourse {R}elations.
\newblock \emph{D\&D}, 4(2):249--281.

\bibitem[{Taboada and Lavid(2003)}]{taboada2003rhetorical}
Maite Taboada and Julia Lavid. 2003.
\newblock Rhetorical and {T}hematic {P}atterns in {S}cheduling {D}ialogues: A
  {G}eneric {C}haracterization.
\newblock \emph{Functions of Language}, 10(2):147--178.

\bibitem[{Taboada and Mann(2006)}]{taboada2006applications}
Maite Taboada and William~C Mann. 2006.
\newblock Applications of {R}hetorical {S}tructure {T}heory.
\newblock \emph{Discourse Studies}, 8(4):567--588.

\bibitem[{Zeldes(2016)}]{Zeldes2016}
Amir Zeldes. 2016.
\newblock rst{W}eb - {A Browser-based {A}nnotation {I}nterface for {R}hetorical
  {S}tructure {T}heory and {D}iscourse {R}elations}.
\newblock In \emph{Proceedings of NAACL-HLT 2016 System Demonstrations}, pages
  1--5, San Diego, CA.

\bibitem[{Zeldes(2017)}]{zeldes2017gum}
Amir Zeldes. 2017.
\newblock The {GUM} {C}orpus: {C}reating {M}ultilayer {R}esources in the
  {C}lassroom.
\newblock \emph{Language Resources and Evaluation}, 51(3):581--612.

\bibitem[{Zeldes(2018{\natexlab{a}})}]{Zeldes2018}
Amir Zeldes. 2018{\natexlab{a}}.
\newblock \emph{Multilayer Corpus Studies}.
\newblock Routledge Advances in Corpus Linguistics 22. Routledge, London.

\bibitem[{Zeldes(2018{\natexlab{b}})}]{zeldes2018neural}
Amir Zeldes. 2018{\natexlab{b}}.
\newblock A {N}eural {A}pproach to {D}iscourse {R}elation {S}ignaling.
\newblock \emph{Georgetown University Round Table (GURT) 2018: Approaches to
  Discourse}.

\end{thebibliography}
\bibliographystyle{acl_natbib}

\end{document}